# Logical settings for concept learning from incomplete examples in First Order Logic.

D. Bouthinon[2,] H. Soldano[2,3] V. Ventos[1],

[2]Laboratoire d'Informatique de l'université Paris Nord (LIPN), UMR CNRS 7030
Av. J-B Clément, 93430 Villetaneuse (France)
[3]Atelier de BioInformatique, Dept des sciences de la vie, Université
Paris VI, 12 rue cuvier 75005 Paris (France)
[1]Laboratoire de Recherche en Informatique de l'Université Paris SUD (LRI),
Bat 490, 91408 Orsay (France)

**Abstract.** We investigate here concept learning from incomplete examples. Our first purpose is to discuss to what extent logical learning settings have to be modified in order to cope with data incompleteness. More precisely we are interested in extending the learning from interpretations setting introduced by L. De Raedt that extends to relational representations the classical propositional (or attribute-value) concept learning from examples framework. We are inspired here by ideas presented by H. Hirsh in a work extending the Version space inductive paradigm to incomplete data. H. Hirsh proposes to slightly modify the notion of solution when dealing with incomplete examples: a solution has to be a hypothesis *compatible* with all pieces of information concerning the examples. We identify two main classes of incompleteness. First, uncertainty deals with our state of knowledge concerning an example. Second, generalization (or abstraction) deals with what part of the description of the example is sufficient for the learning purpose. These two main sources of incompleteness can be mixed up when only part of the useful information is known. We discuss a general learning setting, referred to as "learning from possibilities" that formalizes these ideas, then we present a more specific learning setting, referred to as "assumption-based learning" that cope with examples which uncertainty can be reduced when considering contextual information outside of the proper description of the examples. Assumption-based learning is illustrated on a recent work concerning the prediction of a consensus secondary structure common to a set of RNA sequences.

## 1 Introduction

This work finds its origins in a molecular biology problem: the search for a secondary substructure found on most molecules of a set of RNA sequences. In this problem the data presented an extreme form of incompleteness: the structural information about the sequences is ambiguous as for each sequence many structures could be the real one. A framework addressing such a situation has been presented in [Bouthinon and Soldano, 1998]. This first work led us to a more general investigation of incompleteness of examples in concept learning frameworks, and more precisely, we investigate here the effect of incompleteness on the logical settings for concept learning. We do not propose any concept learning algorithm, however we address various theoretical and practical issues concerning such algorithms.

Usually concept learning from examples relies on a membership relation between hypotheses and examples denoted as *cover* and such that to be a solution an hypothesis has to *cover* positive examples and should *not cover* negative examples of the target concept [De Raedt, 1997]. This set of solutions, inheriting its partial order from the Hypothesis language is called the Version Space [Mitchell, 1982] of the learning problem. This definition of concept learning relies on a complete description of the examples. In [Hirsh, 1994], the author proposes to slightly extend the notion of solution in order to use any piece of information concerning the current example set. In this view, the definition of concept learning problems has also to be modified: a hypothesis has now to be *compatible* with any piece of information related to an example of the current learning set.

Our central interrogation, following Hirsh, is what are the consequences of incompleteness in examples on the solutions of a concept learning problem? We argue hereafter that there exists two basic kind of incompleteness: abstraction and uncertainty that can also be mixed up. The definition of *compatibility* will then depend on the nature of incompleteness of examples. We will generically say that a candidate hypothesis has to be *compatible* with both positive and negative incomplete examples and we will define *compatibility* as a pair of relations *(compatible$^+$, compatible$^-$)* in such a way that a candidate hypothesis has to be *compatible$^+$* with positive examples and *compatible$^-$* with negative examples. In order to illustrate these notions, we consider examples of the concept *fly*. Each example is a bird that can be described with the following features: it is *light* or *not-light*, it is *migratory* or *not-migratory*, and it has one and only one color (either *red* or *green*). Here a hypothesis covers a complete example if all features of the hypothesis belong to the example.

Pure abstraction leads to *generalized examples. A generalized example* is represented by only part of the complete description of the example. This part is supposed to represent a *sufficient* amount of information to determine the label of the example. As a consequence such an example corresponds to a set of complete examples. A hypothesis is then *compatible$^+$* with a positive generalized example if it covers *all* the corresponding positive complete examples. Let us consider a *green light migratory* bird, considered as a positive example. Let us suppose that concerning this particular bird we know that its label (i.e. positive) does not depend on its color, thus resulting in the generalized example *light migratory*. The hypothesis *light or red* is *compatible$^+$* with our example as the hypothesis covers both completed examples *green light migratory* and *red light migratory*. In the same way a hypothesis is *compatible$^-$* with a negative generalized example if it covers *none* of the corresponding negative complete examples. Here *light or red* is not *compatible$^-$* with the generalized negative example *not-light migratory* as *light or red* does cover one of the corresponding complete examples, namely *red not-light not-migratory*. In this last case the hypothesis is correctly rejected though complete description of our negative example (that includes *green*) would fail to reject the hypothesis. Note that here *compatible+* is not the negation of *compatible$^-$* and so we do need a pair of relations rather than a unique coverage relation.

Uncertainty appears when the values of some features are unknown (and cannot be determined) when describing an example. Whenever uncertainty concerns the part of the description that would be sufficient to determine the label of the example, we have to deal with both abstraction and uncertainty. An *uncertain generalized* example is then represented as a set of <u>possible</u> generalized examples. One of these generalized examples corresponds to our observed example. A hypothesis is then *compatible* with such an example if it is *compatible* with at least one possible generalized example. Let us consider again the two examples mentioned above but consider that we do not know whether they are *light* or *not-light*. Our positive example is now represented as the two possible generalized examples *light migratory* and *not-light migratory*. When checking the hypothesis stating that *light or red* birds fly we observe that the hypothesis is *compatible$^+$* with the first possible generalized example *light migratory*. As a consequence, *light or red* is *compatible$^+$* with our uncertain generalized example *{{light, migratory}, {not-light, migratory}}*. Our negative example is now represented as the same uncertain generalized example *{{light, migratory}, {not-light, migratory}}*. Our hypothesis is also *compatible$^-$* with the second possible generalized example

*{not-light, migratory}* and as a consequence the hypothesis is compatible with our negative uncertain generalized example.

Our first purpose here is to define a logical learning setting to deal with incomplete examples. We start from the *Learning from interpretations* learning setting defined in [De Raedt and Dehaspe, 1997a]. *Learning from interpretations* extends to first order languages the basic propositional learning setting to which most attribute-value concept learning methods are related. We then define *Learning from Possibilities* that extends *learning from interpretations* to cope with uncertain generalized examples illustrated above. In [De Raedt, 1997], the author defines *Learning from satisfiability* and shows that the main logical learning settings, as *learning from interpretations* and *learning by entailment*, reduce to (i.e. may be seen, after applying a transformation, as special cases of) *Learning from satisfiability*. We show here that *Learning from Satisfiability* in turn reduces to *Learning from Possibilities* and discuss issues related to the class of hypotheses that are investigated.

The second purpose of the paper is to discuss a particular case of learning from Possibilities that we denote as *Assumptions based Learning.* Here the purpose is to use some information about the example that lies outside its description, as designed in a learning task, in order to reduce the uncertainty of the example. We discuss the role of local and background knowledge in Assumptions based Learning and discuss various properties concerning the case in which hypotheses are DNF formulas containing only positive literals (for instance $H = \exists xy \, (cube(x) \wedge Near(x,y)) \vee \exists x \, (sphere(x)) \vee ....$ ). We briefly discuss our solution [Bouthinon and Soldano, 1999] to a RNA secondary structure prediction problem as an illustration of Assumptions based Learning, and we discuss practical issues related to this real-world problem.

Finally we discuss previous work addressing data incompleteness. In particular we discuss handling of Missing Values in propositional concept learning methods, and integration of abduction and induction in Inductive Logic Programming.

## 2 Preliminaries

### 2.1 Learning and classification of incomplete examples

In the Version space view of noise free concept learning from examples, as proposed by T. Mitchell [Mitchell, 1982], each example, either positive or negative, acts as a constraint that eliminates some hypotheses from the concept space. The version space *Sol(E)* is then the set of solutions of the concept learning task, given the current set *E* of positive and negative examples. Following Hirsh [Hirsh, 1994] we argue that such a constraint can represent any piece of information that shrinks the version space. In our view a (possibly incomplete) example precisely is such a piece of information extracted from some observation. To express concept learning in this more general view we use the following notion of compatibility that relates hypotheses to examples:

Definition: *A hypothesis H is compatible with an example whenever this example does not eliminate the hypothesis H from the Version Space.*

As a consequence, the set of solutions (i.e. the Version space) of a noise free concept learning problem is defined as the set of hypotheses that are *compatible* with all the (possibly incomplete) positive and negative examples. However the definition of compatibility will depend on the class (either positive or negative) of the example. So compatibility will be represented as a pair of relations *(compatible$^+$, compatible$^-$)* relating hypotheses to examples. Usually concept learning from examples [De Raedt, 1997] relies on a membership relation between hypotheses and examples denoted as *Cover* and such that a hypothesis should cover positive examples and should *not* cover negative examples. This requires that *compatible$^-$* is defined as *not compatible$^+$*. However we will further see that this constraint has to be relaxed when dealing with incomplete examples.

The basic property of the Version space is that hypotheses are monotonically rejected when some new complete example *e* is provided. This basic property that holds for complete examples should still hold for incomplete examples. We consider here that some new constraint *c* on the learning set is provided thus turning E in *E'= E ∪{c}*. Such a constraint corresponds either to a new (possibly incomplete) example or to some new information reducing uncertainty on a previous example, or generalizing some previous example. Anyway *Monotonicity of Elimination* has to be preserved:

Definition (Monotonicity of Elimination):
  *A hypothesis eliminated from the version space corresponding to a learning set E will stay outside the Version Space whatever new constraint c is added to the learning set E.*

To further discuss concept learning from incomplete examples, we will use a generic learning setting definition relying on the compatibility relation. The compatibility relation will depend on the nature of incompleteness of examples:

Definition : (Generic Learning setting):
*H is a solution to the learning problem associated to the learning set E divided into positive examples $E^+$ and negative examples $E^-$ iff*
   *H is compatible$^+$ with all examples in $E^+$, and*
   *H is compatible$^-$ with all examples in $E^-$*

When learning is achieved and that one hypothesis *H* is selected, new instances should be classified as either positive or negative. In order to classify a new incomplete instance we have to check whether positive and negative classification of the instance are *consistent* with the selected hypothesis *H*. When dealing with incomplete examples various situations may occur: positive and negative classification being both consistent or none of them being consistent with the selected hypothesis. Consistency will be expressed using the compatibility relations:

Definition: (Consistency):
- *Positive classification of an instance i is consistent (with the selected hypothesis H) whenever H is compatible$^+$ with the instance.*
- *Negative classification of an instance i is consistent (with the selected hypothesis H) whenever H is compatible$^-$ with the instance.*

This leads to the following generic classification setting:

Definition (Generic classification setting):
*A new instance is classified as*
- *Positive whenever positive classification is consistent and negative classification is not.*

*- Negative whenever negative classification is consistent and positive classification is not.*
*- Uncertain, whenever both positive and negative classification are consistent.*
*- Contradictory, whenever neither positive nor negative classification is consistent.*

As we will further see, uncertain classification happens when instances are uncertain, and contradictory classification happens when instances are abstracted.

In what follows we consider as hypotheses, formulas of a logical language. Constraints on a hypothesis concerns which interpretations should be or not models of the hypothesis.

## 2.2  Logic

In this paper we focus on Inductive Logic Programming frameworks [Flach, 1992; Muggleton, 1992; De Raedt 1997] where examples and hypothesis are formalized in logical languages. We present in hereunder syntactical and semantic elements of logic (see [Nienhuys-Cheng and Wolf, 1997] chapters 2 and 3 for details) useful to describe the logical learning settings presented hereafter. All these settings are based on a first-order language $L$ whose syntax and semantics are described below. Note that the reader is supposed to be familiar with basic foundations of logics.

### 2.2.1  Syntax

From a syntactic point of view $L$ is based on a set of predicates $P$, a set of variables $V$, no functional symbols other than a set of constants $HU$, the logical relations $\{\wedge, \vee, \leftarrow\}$ and the quantifiers $\{\exists, \forall\}$. A term $t$ of $L$ is either a constant or a variable. An atom $p(t_1,..,t_k)$ is composed of a predicate symbol $p$ and $k$ terms ($k \geq 0$). A literal is either an atom, (then denoted as a positive literal), or the negation $\neg a$ of an atom $a$ (then denoted as a negative literal). A *clause* $a_1 \vee ... \vee a_m \vee \neg b_1 \vee ... \vee \neg b_n$ is a disjunction of literals ($a_i$ and $b_j$ are atoms), and may be written as $a_1 \vee ... \vee a_m \leftarrow b_1 \wedge ... \wedge b_n$. A *Horn* clause is either written as $a \leftarrow b_1 \wedge ... \wedge b_n$ (a definite clause) or as $\leftarrow b_1 \wedge ... \wedge b_n$ (a negative clause). Note that as literals are clauses, a positive literal $a$ will often be written $a \leftarrow$, and a negative literal $\neg a$ will be written $a \leftarrow$. In this paper we only consider universally quantified clauses, i.e. clauses in which all variables appearing in a clause are in the scope of a universal quantifier $\forall$. For instance the clause $\forall x \forall y (p(x) \leftarrow q(x,y))$ will be written $\forall (p(x) \leftarrow q(x,y))$ or even $p(x) \leftarrow q(x,y)$. A clausal theory is a conjunction of universally quantified clauses $\forall c_1 \wedge ... \wedge \forall c_q$ (each $c_i$ is a clause), which will be also denoted as a set of clauses $\{c_1; ...;c_q\}$. We also call CNF formulas such clausal theories.

Let us define a *cube* as a conjunction of literals $\neg a_1 \wedge ... \wedge \neg a_m \wedge b_1 \wedge ... \wedge b_n$ (where $a_i$ and $b_j$ are atoms). We will call DNF formula a disjunction of existentially quantified cubes as $\exists d_1 \vee ... \vee \exists d_q$ (where each $d_i$ is a cube). For instance $\exists x \exists y (\neg p(x) \wedge q(y)) \vee \exists z\, r(z)$ is a DNF formula. A DNF$^+$ formula is a DNF formula with no negative literal.

Note that a cube is the negation of a clause (and conversely), and a DNF formula is the negation of a CNF formula (and conversely).

Any CNF or DNF formula (and more specifically a clause, a cube or a literal) is said *ground* when it contains no variables (in such a case quantifiers are irrelevant).

### 2.2.2  Semantics

The semantics says how to assign a truth value (either TRUE or FALSE) to a formula.

In this paper we will focus on Herbrand interpretations that are based on a specific domain of interpretation, called a Herbrand universe and that simply is the set of constants $HU$. $HU$ is the semantic domain used to interpret terms (constant and variables): a constant $c$ of $HU$ is interpreted as itself, and any variable $v$ is interpreted as an element $e$ of $HU$ thanks to the substitution $\{v/e\}$. Thus given a non quantified formula $f$ having $q$ variables $\{v_1, ..., v_q\}$, $f\theta$ is a ground formula iff $\theta = \{v_1/e_1, ..., v_q/e_q\}$ substitutes each occurrence of each variable $v_k$ in $f$ with the element $e_k$ of $HU$. For instance consider $HU=\{a,b\}$, then $light(a) \leftarrow red(a)$ is a ground formula built from $light(x) \leftarrow red(x)$ applying the substitution $\theta = \{x/a\}$.

Let us define the Herbrand base $HB$ as the set of ground atoms that can be formed from the predicates in $P$ and the constants in $HU$. A Herbrand interpretation $i$ assigns a truth value to each ground atom of $HB$. $i$ divides $HB$ into the set $i_p$ of true ground atoms (with truth value TRUE) and the set $i_n$ containing false ones (with truth value FALSE). For instance consider $HU = \{a,b\}$ and $P = \{light, red, green\}$ in which each predicate is unary, then $HB = \{light(a), light(b), red(a), red(b), green(a), green(b)\}$. Then a Herbrand interpretation $i$ is defined as $i_p = \{light(a), light(b), red(a), green(b)\}$ and $i_n = \{red(b), green(a)\}$. Note that from $HB$ and $i_p$ we can infer $i_n$, so when $HB$ is fixed, the interpretation $i$ is simply represented as its true part $i_p$.

We can now assign a truth value to any DNF or CNF formula from an Herbrand interpretation $i$ based on the Herbrand base $HB$ :

- a ground atom $a$ is TRUE under $i$ iff $a$ is in $i_p$
- a ground negative literal $\neg a$ is TRUE under $i$ iff $a$ is not in $i_p$
- a universally quantified clause $\forall (l_1 \vee ... \vee l_m)$ (each $l_k$ is a literal) is TRUE under $i$ iff for each substitution $\theta$ grounding $l_1 \vee ... \vee l_m$ at least one ground literal $l_k\theta$ is TRUE under $i$
- a CNF formula $\forall c_1 \wedge ... \wedge \forall c_q$ is TRUE under $i$ iff each clause $\forall c_i$ is TRUE under $i$
- an existentially quantified cube $\exists (l_1 \wedge ... \wedge l_m)$ is TRUE under $i$ iff it exists a substitution $\theta$ grounding $(l_1 \wedge ... \wedge l_m)$ in such a way that each ground literal $l_k\theta$ is TRUE under $i$
- a DNF formula $\exists d_1 \vee ... \vee \exists d_q$ is true under $i$ iff at least one cube $\exists d_j$ is TRUE under $i$.
- Any DNF or CNF formula that cannot be computed as TRUE under $i$ is FALSE under $i$.

Consider for instance the Herbrand interpretation $i$ defined above. The existentially quantified cube $c= \exists x (light(x) \wedge green(x))$ is true under $i$ since the substitution $\theta = \{x/b\}$ grounds $c$ and $light(b)$ and $green(b)$ are true under $i$.

A Herbrand interpretation $i$ built on $HB$ and under which a formula $f$ is true is called a Herbrand *model* of $f$. We will further use the following notation:

Consider two formulas $f$ and $g$ :
$f \mid\neq_{HB} \square$ means that $f$ has at least one Herbrand model,
$f \mid=_{HB} \square$ means that $f$ has no Herbrand model,
$f \mid=_{HB} g$ means that each Herbrand model of $f$ is an Herbrand model of $g$.

Note that all these relations depend on the Herbrand Base $HB$. However when the Herbrand base is not ambiguous, we use the notation $\mid=$ instead of $\mid=_{HB}$, and we say interpretation instead of Herbrand interpretation.

Furthermore, in this paper we will often define the language (i.e. the predicates $P$ and the constants $HU$) starting from a given Herbrand base $HB$. So $P$ and $HU$ are the sets of predicate and constant symbols appearing in $HB$.

## 3 Concept learning from interpretations.

We discuss here data incompleteness in the learning from interpretations setting as defined by L. De Raedt. This learning setting extends to first order logics, the basic propositional learning setting to which most attribute-value concept learning methods are related. In the propositional learning setting an example is an interpretation, i.e. it is the assignment of truth-values to the set of atomic propositions of a propositional language. De Raedt considers interpretations built on a Herbrand Base, thus extending this learning setting to first order clausal theories. As seen in §2.2.2 an example corresponds then to the assignment of truth-values to the set of grounded atoms built from a first order language. A similar learning framework is found in [Khardon, 1999] with as a motivation the extension of learnability results and methods to first order languages.

In concept learning from interpretations, there are basically two classes of hypotheses in investigation:
- The hypothesis corresponds to a theory i.e. a conjunction of facts and rules. Each rule holds on all positive examples, and each negative example should contradict the theory. For instance each clause in H= {square(X) ← light(X); light(X) ←white(X)} holds on the two positive examples *{square(a), light(a)}* and *{square(a),white(a), light(a)}*, and H contradicts the negative example *{light(a)}* as H implies that any *light* object should also be *square*. The theory is then a CNF formula. CLAUDIEN [De Raedt and Dehaspe, 1997], and, in the case of attribute-value representations, CHARADE [Ganascia, 1993] are examples of this first learning situation often referred to as "descriptive„ learning, and most often only positive examples are used. However systems as ICL [Van Laer *et al*, 1997] (using its CNF option) or LOGAN_H [Khardon, 2000] use both positive and negative examples and build a clausal theory whose purpose is to classify new examples.
- The hypothesis corresponds to a concept definition, i.e. the disjunction of various partial definitions $h_i$. Each partial definition covers part of the positive examples and no negative examples. Consider for instance H= ∃X (square(X)∧light (X)) ∨ ∃X (white(X)∧light (X)), the first term of H covers *{square(a), light(a)}* and the second covers *{white(a), light (a)}* while H does not cover the negative example *{light(a)}*. Note that the concept definition is a DNF formula. The purpose is then clearly to classify new instances. ICL (in its DNF mode) is an example of this second situation.

In what follows we make no assumptions regarding the class of formulas allowed as hypotheses. However within most examples illustrating the ideas presented below, our hypotheses will be DNF formulas. We also will discuss in §4.4 to what extent CNF formulas can be learned using algorithms designed to learn DNF formulas and vice versa.

### 3.1 Learning from complete examples

Hereafter we will suppose that examples and hypothesis are built from a set of predicates $P$, and that a specific domain $HU_k$, i.e. a specific Herbrand universe representing observed entities, is associated with each example. More precisely the example $i_k$ is a Herbrand interpretation based on the Herbrand base $HB_k$ that is built from $HU_k$ and $P$.

*Example 1*

> Let *P={light(.), red(.), green(.), brighter(.,.)}* be a set of predicates.
> Our first example is made of two light objects so : $HU_1$ = *{a, b}* and as a consequence $HB_1$= {*light(a), light(b), red(a), red(b), green(a), green(b), brighter(a,a), brighter(a,b), brighter(b,a), brighter(b,b)*}. We know that the red one is brighter than the green one. In consequence the example is defined by the following Herbrand interpretation built on $HB_1$ :

$i_{p1}$ = { light(a), light(b), red(a), green(b), brighter(a,b)}

Our second example is made of three objects *a*, *b* and *c*, where *a* and *b* are light objects but *c* is not light, and only *a* and *b* have a color:

$HU_2$ = {a, b, c} and $i_{p2}$ = { light(a), light(b), red(a), green(b), brighter(a,b)}.

The two examples are different: though they have the same positive part, in the second example *green(c)* is false whereas in the first example *green(c)* is meaningless, as *c* does not belong to the universe describing this example.

The hypothesis space is made of either DNF formulas or CNF formulas built from *P*, a set of variable *V*, and a set (possibly empty) of constants $HU = \cap HU_i$. This means that a constant that appears in a hypothesis has to belong to the domain of each example *i* (and further instances). Such constants (as for instance, numbers 1, 2, 3, ...) should be considered as "universal" constants with similar properties in all examples. For sake of clarity, we further avoid constants within hypotheses in our illustrations. [Khardon, 2000] proposes a similar representation though the author does not explicitly use Herbrand interpretations.

Compatibility is defined in Learning from interpretation as $compatible_I=(compatible_I^+, compatible_I^-)$ as follows:

<u>Definition</u>: *H is $compatible_I^+$ with the positive example i iff i is a model of H, and H is $compatible_I^-$ with a negative example i iff ī is not a model of H.*

Note that $compatible_I^-$ is defined as *not $compatible_I^+$* : a unique relation *Cover* defined as "*H* covers the example *i* iff *i* is a model of *H*" is sufficient to define the solutions *H* in the usual way:

    *H* is a solution iff:

    *H* covers all positive examples and no negative example.

This relation is illustrated in the following Figure 1 [1]

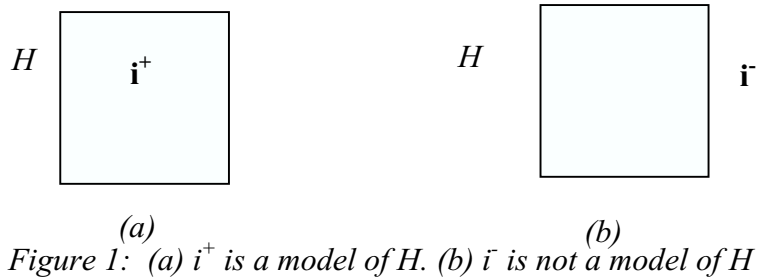

*Figure 1:* (a) $i^+$ is a model of H. (b) $ī^-$ is not a model of H

In example1, the clause $\forall x\ light(x)$ ( "all objects of the example are light ") covers the first example and does not cover the second one.

Classifying a new (complete) instance w.r.t the selected definition *H* is straightforward in learning from interpretations: either positive classification of *i* is consistent (*i* is a model of

---

[1] We will figure the set of models *M(H)* of a hypothesis *H* as a rectangle, and the set of models *M(e)* of a formula *e* as an ellipse, moreover we will represent an interpretation by its name in bold characters.

$H$) or negative classification is consistent ($i$ is not a model of $H$). As a result a new instance $i$ is classified either as positive or as negative.

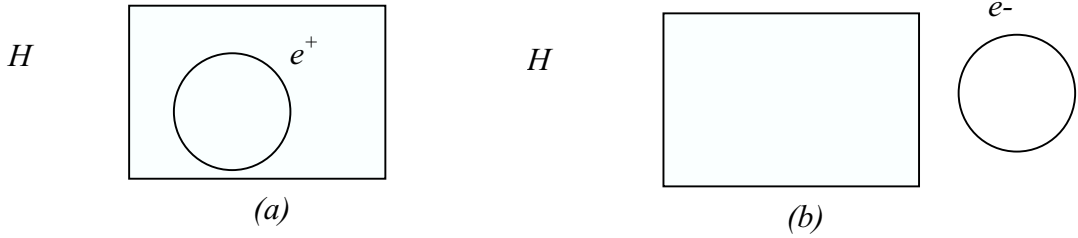

Figure 2: (a) $H$ is compatible$_G^+$ with $e^+$: each model of $e^+$ is a model of $H$
(b) $H$ is compatible$_G^-$ with $e^-$: no model of $e^-$ is a model of $H$

In order to extend concept learning from interpretations, we will both change the representation of the examples, and instantiate the compatibility relations.

### 3.2 Learning from Generalized examples

A constraint stronger than one corresponding to a complete example is obtained by generalizing the example. A generalized example is a clausal theory $e$ to which is associated a Herbrand base $HB$. Each model built on HB of $e$ acts on the solution space as a complete example. Generalized examples express the idea that only part of the (complete) observed example is sufficient to state that the example satisfies (or does not satisfies) the target concept. Let us suppose, for instance, that we learn the target-concept *fly* about birds using various attributes including color attributes, and suppose that when dealing with a *migratory bird,* the color is unrelated to the target concept. This means that whether we encounter a migratory bird as a positive example we have to describe it as a generalized positive example. More precisely all instances whose differences from the positive example only concern colors, should also be covered by the definition of the target concept. In the same way, if our migratory bird represents a negative example of our target concept, then we will describe it as a generalized negative example, and, as a consequence, no example identical to the negative example, except concerning its color, should be covered by the definition of the target concept.
Thus the compatibility relation pair *compatible$_G$* is defined as follows:

Definition
- $H$ is compatible$_G^+$ with a generalized positive example $e^+$ associated to the base HB iff each model of $e^+$ built on HB is a model of $H$, i.e. :
$$e^+ \models_{HB} H.$$
- $H$ is compatible$_G^-$ with a generalized negative example $e^-$ associated to the base HB iff no model of e- built on HB is a model of $H$, i.e. :
$$e^- \wedge H \models_{HB} \square$$

As illustrated in Figure 2, the positive constraint means that $H$ covers each complete example within $e^+$, whereas the negative constraint means that $H$ does not cover any complete example within $e^-$.

*Example 2*
Suppose we have a propositional language defined by the atomic propositions $P = \{bird, light, red, green\}$

Let $e^+$ = {bird; light)} be the clausal theory representing a generalized positive example. As we do not state the color of our example, $e^+$ has 4 models:
$i_{p1}$ ={ bird, light}, $i_{p2}$ = { bird, light, <u>red</u>}, $i_{p3}$ ={ bird, light, <u>green</u>}, $i_{p4}$ ={ bird, light, <u>red</u>, <u>green</u>},
the hypothesis $H$ = (bird∧light) ∨ (red∧light) is compatible$^+$ with $e^+$ (all models of $e^+$ are models of $H$).
Let $e^-$ = {bird ; ← light} represent a generalized negative example. Here again we do not state the color of *the object,* therefore $e^-$ has again 4 models:
$i'_{p1}$ = { bird}, $i'_{p2}$ = { bird, <u>red</u>}, $i'_{p3}$ = { bird, <u>green</u>}, $i'_{p4}$ = { bird, <u>red</u>, <u>green</u>}
and $H$ is also compatible with $e^-$ (no model of $e^-$ is a model of $H$).
Note that the literal *red* appears in $H$, possibly allowing $H$ to be compatible with some other positive example, in which color is relevant.

A generalized example represents a set of complete examples that all have to be compatible with the hypothesis. As a consequence Monotonicity of Elimination is inherited from the case of complete examples.

Learning and classification settings are obtained from our generic learning setting using the compatibility relation pairs defined above. Now considering generalized instances, what is the intended meaning of a generalized instance? Suppose that concerning migratory birds we know that their class does not depend on their color. This means that the selected concept definition should *classify* such a migratory bird, thus considered as a generalized instance, in the same class whatever we know about its color.

Here *contradictory* classification appears whenever neither positive classification nor negative classification is *consistent* with our concept definition $H$. As a consequence our concept definition is no more satisfactory.
In our previous example suppose that we selected the hypothesis
   $H$ = bird ∧ red, and that we have to classify the generalized instance
   $e$ = {bird; migratory};
the generalized instance is classified as *contradictory:* we intended that the classification of our instance should not depend on the color of migratory birds, and it does depend on their color.

In other words generalized examples offers a way to use unlabeled examples to reject a hypothesis, and so to shrink the solution space.

Finally, we should mention that a generalized example is not always related to one observation, and so to one particular Herbrand Base, but may represent some piece of background knowledge . For instance the following generalized positive example:
   $e$= { ∀bird(X)}
means that whenever all entities in an instance are birds then it should be classified as positive.

## 4 A general setting to learn from incomplete examples

### 4.1 Learning from Possibilities

We have seen in §3.2 that a clausal theory together with a Herbrand Base represent what is sufficient to state about an example for relating it to a target concept.

A more general case arises when we face both uncertainty and generalization, i.e. when we don't know all that would be sufficient to relate our example to the target concept. The most general situation we can think of has the two following properties: 1) uncertainty in our observation is such that we do not know which description amongst several possible descriptions correspond to our observation, and 2) each description represents what we should know about our observation for our learning purpose.

Imagine for instance that we observe a robot, which is a positive example of the concept *Task1*. Also suppose that the arm of the robot is hidden and that either the arm holds a yellow hammer or it holds a screwdriver and a screw. Furthermore we know that in the later case the color of the screwdriver does not matter as we do not need any more information to consider the robot as a positive example of Task1. So our example $e^+$ will be represented by two *possible* generalized examples $P_1$ and $P_2$. Both $P_1$ and $P_2$ include a set of clauses $B$ that represents background knowledge asserting, for instance, that the same object cannot be both a hammer and a screwdriver:

$e^+ = \{ P_1, P_2\}$
$P_1 = \{arm(a); Holds(a,b); Hammer(b); Yellow(b); B\}$ with $HU=\{a,b\}$
$P_2 = \{arm(a); Holds(a,b); Screwdriver(b); Holds(a,c); Screw(c); B\}$ with $HU=\{a,b,c\}$

Note that the subdomain defining the description of the robot depends on the "possibility" $P$ under consideration (i.e. $HU(P_1) = \{a,b\}$ and $HU(P_2) = \{a,b,c\}$)

We can think of $P_1$ as a complete example (a particular case of generalized example) in which the arm of the robot only holds a hammer. Hence only constants $a$ and $b$ are used to describe the example. $P_2$ is a generalized example, as we do not mention the color of the screwdriver (it could be yellow, but it does not matter).

To be compatible with our uncertain-generalized example, a hypothesis has to be compatible with $P_1$ or compatible with $P_2$. For instance $H = \exists XY\ Holds(X,Y) \land Screw(Y)$ is compatible$^+$ with $e^+$ as $H$ is compatible$_G^+$ with $P_2$. In other words, $H$ is compatible with the positive example because, under the assumption that the hidden arm of the robot holds a screwdriver and a screw, asserting $P_2$ is sufficient to state that $H$ is satisfied, and so $H$ is an acceptable definition for the target concept.

So we propose to represent an uncertain generalized example as a set $e=\{P_1, ...P_n\}$ of consistent formulas, where each $P_i$ stands for a "possible" generalized example.
In Learning from Possibilities, the compatibility relation pair *compatible$_P$* is so defined as follows:

<u>Definition</u> *(Compatibility in learning from Possibilities)*:
 -$H$ is compatible$_P^+$ with the example $e^+ = \{P_1, ...P_n\}$ iff
  $\exists P_i \in e^+$ such that $P_i \models_{HB} H$
 -$H$ is compatible$_P^-$ with the example $e^- = \{P_1, ...P_n\}$ iff

$\exists P_i \in e^-$ such that $P_i \wedge H \models_{HB} \square$

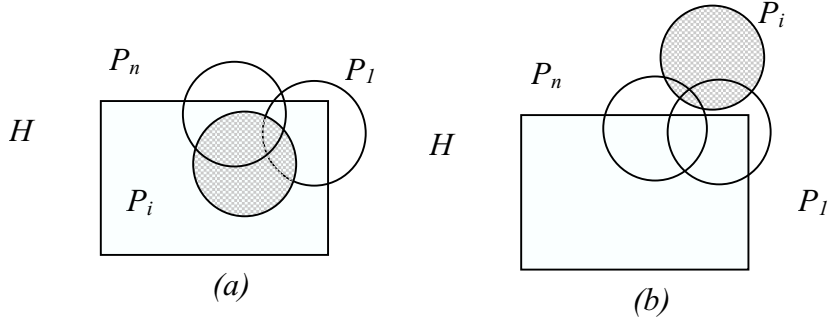

*Figure 3* (a) $H$ is $compatible_P^+$ with $e^+$ (b) $H$ is $compatible_P^-$ with $e^-$

In other words to be $compatible_P^+$ with a positive uncertain generalized example $e^+$, a hypothesis $H$ has to be $compatible_G^+$ with at least one of the possibilities of $e^+$. In the same way, to be $compatible_P^-$ with a negative uncertain generalized example $e^-$, a hypothesis $H$ has to be $compatible_G^-$ with at least one of the possibilities of $e^-$. These relations are illustrated in Figure 3.

*Monotonicity of Elimination* is preserved in Learning from Possibilities as any new information concerning an uncertain-generalized example $e$ results in eliminating some possibility $P_i$ from $e$.

Regarding classification of uncertain-generalized examples, our generic classification setting depends on consistency of positive and negative classification. *Contradictory* classification occurs when there is neither a possibility $P_i$ the models of which are all models of $H$, nor a possibility $P_j$ of $e$ that contains no models of $H$. *Uncertain* classification occurs *when* there is both a possibility $P_i$ the models of which are all models of $H$, and a possibility $P_j$ of $e$ that contains no models of $H$

Note that pure uncertain examples, generalized examples, and complete examples are particular cases of examples in learning from possibilities*:*

*a) Generalized examples*
Whenever $e = \{P_1\}$ contains one single possibility associated to a Herbrand base, $e$ represents a generalized example as defined in §3.2

*b) Complete examples*
Whenever $e = \{P_1\}$ contains one single possibility $P_i$ that has exactly one model built on the Herbrand base $HB$ associated to $e$, then $e$ represents a complete example as defined in §3.1.

*c) Pure uncertain examples*
Whenever $e = \{P_1, ...P_n\}$ and each possibility $P_i$ has exactly one model built on a unique Herbrand base $HB$ associated to $e$, then $e$ corresponds to a set of interpretations on $HB$. Each of these interpretations refers to one possible *identity* of the observed example. Such a *pure uncertain example* often corresponds to a formula of $L$ whose models on $HB$ are enumerated in $e$. If we also denote as $e$ this formula we obtain the following definition of compatibility:
<u>Definition</u>:
-$H$ is $compatible_U^+$ with a pure uncertain example $e^+$ if there is some model of $e^+$ on $HB$ that is a model of $H$. This may be rewritten as:
$\quad H \wedge e^+ \not\models \square$

-H is compatible$_U^-$ with a pure uncertain example $e^-$ if there is some model of $e^-$ on HB that is <u>not</u> a model of H. This may be rewritten as:
$e^- \not\models H$

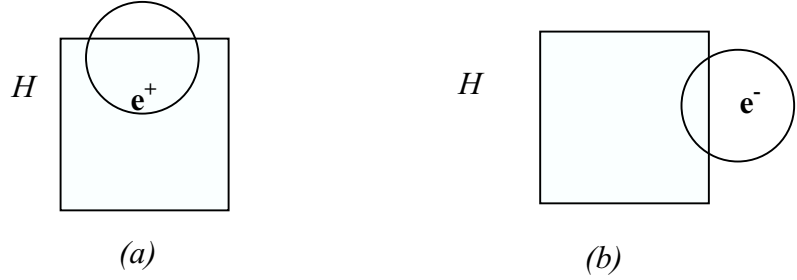

Figure 4: (a) H is compatibl$_U^+$ with $e^+$  (b) H is compatible$_U^-$ with $e^-$

Because uncertain generalized examples can represent uncertain examples, generalized examples or complete examples learning from possibilities is the most general setting presented here.

### 4.2  Learning from Possibilities vs. Learning from satisfiability

By generalizing the representation of an example from a complete interpretation to a clausal theory, [De Raedt, 1997] and [De Raedt and Dehaspe, 1997] define "learning from satisfiability".

The basic motivation of representing an example as a formula (or the set of its models) was to address incompleteness. However in learning from satisfiability this incompleteness does not have the same meaning depending on whether the example is positive or negative.

$E^+$ and $E^-$ are sets of clausal theories. Any accepted hypothesis H is a clausal theory that satisfies:
$e^+ \wedge H \not\models \Box$ for all examples $e^+$ of $E^+$ and
$e^- \wedge H \models \Box$ for all examples $e^-$ of $E^-$.

When referring to our classification of examples, the first equation means that the clausal theory $e^+$ represents a pure uncertain positive example and the second equation means that the clausal theory $e^-$ represent a generalized negative example. This asymmetry technically comes from the formulation of concept learning using a single "cover" relation that is negated when considering negative examples. That's the reason why we formulate concept learning from incomplete data using a pair of compatibility relations. However, consider that satisfiability was originally proposed with two motivations: first to extend learning from interpretations to deal with uncertainty for descriptive learning tasks, second, to reformulate most Inductive Logic Programming learning settings. Concerning the former motivation, as usually no negative examples are used in descriptive learning systems (as in CLAUDIEN) there is no drawback in ignoring uncertainty in negative examples. In the other hand, the ILP learning settings that are reformulated are all related to the so-called "ILP normal setting" whose purpose is to derive positive examples and avoid deriving negative examples. Again a single "cover" relation is sufficient to reformulate these learning settings in which basically all the useful knowledge about examples is supposed to be known.

Another point in which our framework differs from learning from satisfiability, is that Learning from satisfiability represents hypotheses as clausal theories when one of our practical purpose is to extend DNF as found in propositional learning to existential first order DNF. Finally we should note that learning from satisfiability does not explicitly discuss the previous association of a Herbrand base to each example. This is because, as far as examples *e* are represented as clausal theories and hypotheses *H* also are clausal theories, then checking whether *e* ∧ *H* has a model can be safely performed by only considering the Herbrand interpretations built from constants and predicates appearing in *e* ∧ *H* [Nienhuys-Cheng and Wolf, 1997]. When considering other formulas, as DNF hypotheses, or considering other equations as *e* |=*H*, this property does not hold, and so it is necessary to clearly define what constants are to be considered when representing our state of knowledge about each example.

### 4.3  Learning from satisfiability reduces to learning from possibilities

In [De Raedt, 1997] the author shows that the main concept learning settings, as *learning from interpretations* and *learning by entailment* reduce to (i.e. may be seen, after applying a transformation, as special cases of) "learning from satisfiability" and that learning from satisfiability does not reduce to these learning settings.

First, let us introduce the notion of reduction as defined in [De Raedt, 1997]. In De Raedt view, a learning problem is defined as a language of hypothesis $L_c$, a language of examples $L_e$, a membership relation *Cover* that specifies how $L_c$ relates to $L_e$, and a set of examples *E* of some target concept. Each example has the form *e* =(*p, Class*) where *p* belongs to $L_e$ and *Class* is either *Positive* or *Negative*. Various learning settings are compared in this work, depending on the representation of examples and the related coverage relation. The solution set *Sol (E)* is the set of hypotheses that each covers all the positive examples and does not cover any negative example. The notion of reduction allows relating the various type of coverage. The basic idea is that if there exists a reduction *ρ* from learning under *CoverA* to Learning under *CoverB*, then learning problems under *CoverA* can be solved using the data transformation *ρ* and algorithms for learning *CoverB*. As a consequence Learning under *CoverB* seems a more general task than learning under *CoverA*.

Our first purpose here is to show that learning from satisfiability reduces to learning from possibilities. However learning from possibilities relies on a pair of compatibility relations (*Compatible$^+$, Compatible$^-$*), rather than on a unique coverage relation, and as a consequence the definition of the solution set has to be slightly modified : The solution set *Sol (E)* is now the set of hypotheses that are *compatible$^+$* with all the positive examples and *compatible$^-$* with all the negative examples. This leads to the following definition of reduction, where *Compatible$_A$* and *Compatible$_B$* are two compatibility relation pairs, associated to the two learning settings *A* and *B*:

Definition: A reduction from learning under *Compatible$_A$* to learning under *Compatible$_B$* is a function *ρ* that maps any learning set $E_A$ (under *Compatible$_A$* ) onto a learning set $E_B$ = {*ρ(e)* | *e* ∈ *E*} (under *Compatible$_B$*) of *B* such that $Sol_A(E_A) = Sol_B(E_B)$.

The solution set defined under the coverage relation *Cover* is equivalent to the solution set defined under the pair of compatibility relations (*Cover, Not_Cover*). As a consequence our reduction notion is more general that the notion proposed by De Raedt and all the reduction results presented in [De Raedt, 97] are true under our notion of reduction. Furthermore we have the following results:

Proposition 1: *Learning from satisfiability reduce to learning from possibilities*
(proof in appendix A.2)
Proposition 2 : *Learning from possibilities does not reduce to learning from satisfiability*

(proof in appendix A.3)

## 4.4 Learning DNF vs. Learning CNF.

DNF and CNF (see part 3) are two usual classes of formulas that are currently investigated in concept learning. Moreover by negating a DNF we obtain a CNF and vice versa. Furthermore learning from interpretations a formula $H$ from $E^+$ and $E^-$ has be shown as equivalent to learning *not-H* from $E^+$ and $E^-$ [Van Laer et al, 1997]. More precisely let us denote as $Sol_A(L, E=E_+,E_-)$ the set of solutions under compatible$_A$ where $L$ is the class of formulas that are admitted as solutions, let *not-E* be the transformation that switch the labels of the elements of $E$, and let *not-L* be the class of formulas obtained by negating formulas of $L$. Then:

<u>Proposition 3</u>: *Let I represents learning from interpretations (i.e. compatible$_I$= (is-a-model, is-not-a-model)) then $Sol_I(L, E) = Sol_I(not-L, not-E)$. [Van Laer, 1997]*

The same proposition is false in learning from satisfiability, but it's true in learning from possibilities:

<u>Proposition 4</u>: *Let P represents learning from possibilities then $Sol_P(L, E) = Sol_P(not-L, not-E)$ .*
Proof: Let $H$ be an element of $L$ that belongs to $Sol_P(L, E=E_+,E_-)$. This means that
1) for any element $e$ of $E^+$ it exists $p \in e$ such that : $p \models H$ and 2) for any element $e$ of $E^-$ it exists $p \in e$ such that $p \wedge H \models \Box$.
*If we switch the labels of elements of $E^+$, then the elements( e, positive) of $E^+$ are transformed as not( e, positive )=(e, negative). We have then $not(E+)=E'^-$ and following 1) for any element $e$ of $E'^-$, it exists $p \in e$ such that $p \models H$ and so $p \wedge \neg H \models \Box$, i.e. all negative elements in $E'^-$ are compatible$_P$- with not-H.*
*In the same way by transforming elements of $E^-$ as not(e, negative )=(e, positive) we obtain $not(E^-)=E'^+$ and following 2) for any element $e$ of $E'^+$, it exists $p \in e$ such that $p \wedge \neg H \models \Box$ and so $p \models \neg H$, i.e. all positive elements in $E'^+$ are compatible$_P$+ with $\neg H$.*

Now Learning from Interpretations and Learning from satisfiability are both particular cases of Learning from Possibilities. More precisely they both correspond to reductions (see §5.3) in which the labels of the examples are unchanged. So we are interested in characterizing the reductions for which proposition 4 is true. A quick look at the proof above shows that the proof holds as far as the compatibility relations have the following property: *for any example $e$, $H$ is Compatible$^+$ with $e$ iff $\neg H$ is Compatible$^-$ with $e$.* As long as a reduction of Learning from Possibilities preserves this property, Proposition 4 holds. This is the case in particular of *Assumptions Based learning* described in the next section.

## 4.5 Local Knowledge and background Knowledge.

From a theoretical point of view both background knowledge and local knowledge concerning a given uncertain-generalized example may be freely included in the representation of the example as a clausal theory. So there is no theoretical issue related to background or local knowledge in Learning from Possibility. However Learning from Possibility is a very general learning setting and we will discuss the role of background and local knowledge on the more specific learning setting presented in the next section.

# 5   Assumptions based learning

We discuss here a specific case (i.e. a reduction) of Learning from Possibilities. Let us suppose that we face a concept learning problem in which each example is uncertain and that we have some information that we can use to shrink uncertainty, but that we do not want to use to learn the target concept. This simply means that our inductive bias defines how each example should be described but that we need some extra information to reduce our uncertainty about the examples. Such an example is then denoted as an *extended uncertain example*.

More formally we associate to an uncertain extended example a clausal theory $e$. The formula $e$ represents our state of knowledge on the example and is associated to an *extended* Herbrand base $HBe$ such that we only consider models of $e$ built on $HBe$ as possible representations of the example. However the intended complete description of the example for our learning task should be an interpretation built on only a part $HB$ of $HB_e$. $HB$ is also a Herbrand base built from less constants and predicates than $HB_e$. $HB$ is denoted as a *subbase* of $HB_e$. So the extended uncertain example is defined as the clausal theory $e$ together with the extended Herbrand base $HB_e$ and with the Herbrand subbase $HB$.

*Example 3.a*

> Let $HB$ = *{light, white, square}* and $HB_e$ = *{light, white, polygon, square}* expressing that the complete description of the positive example $e$ relies on the truth values of *light, white* and *square,* and that we also use the truth value of *polygon* to help reducing the uncertainty concerning the truth value of *light, white* and *square*:
> $e$ = *{light ; polygon←square;  ←polygon∧ white}*
> Here we know that our example can't be a white polygon, so it can't be both light and square. This results in the two following possible descriptions of the example for our learning purpose:
> *{light, white}*
> *{light, square}*

Note that as in learning from possibilities the interpretations built on $HB$ have to assign a truth value to hypotheses. So all predicates (and possibly constants) appearing in a hypothesis have to appear in all Herbrand bases $HB$ associated to extended uncertain examples.

We give hereunder various definitions and notations before formally defining Assumptions Based Learning:

An interpretation $j$ on the subbase $HB$ of $HB_e$ is called a *partial interpretation*: no truth-value is assigned to atoms of $HB_e$-$HB$.

An interpretation $i$ on $HB_e$ is an *extension* of an interpretation $j$ on $HB$ iff $j_p \subseteq i_p$ and $j_n \subseteq i_n$. Conversely $j$ is the *projection* of $i$ on $HB$. We will denote as *ext(j)* the set of all extensions of $j$ on $HB_e$.

An interpretation $j$ built on $HB$ is a *partial model* of a formula $f$ iff at least one interpretation in *ext(j)* is a model of $f$.

Let $j$ be an interpretation built on $HB$ and such that $j_p = \{t_1,..,t_m\}$ and $j_n = \{f_1,...,f_n\}$. Then we note $ct(j) = ct(j_p) \wedge ct(j_n)$ the clausal theory containing the positive clauses $(ct(j_p)= \{t_1;..;t_m\})$ and the negative clauses $(ct(j_n)= \{f_1\leftarrow;...;f_n\leftarrow\})$ representing the true and false ground atoms of $j$. Note that the interpretation $j$ is the single model of $ct(j)$ built on $HB$, and that $ext(j)$ represents the models of $ct(j)$ built on $HB_e$.

Now given an extended example *e*, the possible complete examples are the partial models *j* (built on *HB*) of *e*. The set of possible complete examples is computed by intersecting models of *ct(j)* and models of *e*. In order to define the compatibility of a hypothesis *H* with an extended example *e*, we will then check whether *H* is compatible with at least one possible complete example issued from *e*. This results in the following definition of *compatibility$_A$* when dealing with uncertain extended examples:

Definition
*H* is compatible$_A^+$ with the extended positive example *e$^+$* iff there exists an interpretation *j* built on *HB* such that:
  $ct(j) \wedge e \not\models_{HBe} \square$   and *j* is a model of *H*

*H* is compatible$_A^-$ with the extended negative example *e$^-$* iff there exists an interpretation *j* built on *HB* such that:
  $ct(j) \wedge e \not\models_{HBe} \square$   and *j* is not a model of *H*

Here *ct(j)* represents all the positive and negative literals that are logical consequences of *e* plus a set of assumptions on the truth-value of the remaining atoms of *HB*.
Note that when $HB = HB_e$, extended examples simply are pure uncertain examples as defined in. §2.2.1.

*Example 3-b*

> The hypothesis *H = white ∧ square* is not *compatible$_A^+$* with the positive example *e* since *(see example 3-a)* the only partial models of *e* are *{light, white}* and *{light, square}*, and none of them is a model of *white ∧ square*.

This should not be confused with inductive biases on the hypothesis language. In the previous example we could have considered the example as a pure uncertain example, represented using the base *HB$_e$*, and stating that an inductive bias exclude *polygon* of the hypothesis language. However such an approach is inaccurate if the desired representation *HB* contains fewer constants than the extended representation *HB$_e$* as shown in the following example:

*Example 4*

> We search for an hypothesis built on the unary predicates *P = {light, white, square}*. In order to reduce uncertainty about an example describing a single object *a*, we use a binary predicate *Near* together with information concerning a second object *b*, so we have:
>   *HB   = {light(a), white(a), square(a)}* and
>   *HB$_e$={light(a), white(a), square(a), light(b), white(b), square(b), Near(a,b), Near(b,a), Near(a,a), Near(b,b)}*
> Our extended uncertain example is represented by the following formula:
>   *e={light(a); Near(a,b); ←Near(x,x); ←square(x) ∧ white(x) ∧ Near(x,y) }*
> This means that the object *a* cannot be both *square* and *white* as *a* is near an other object *b*.
> We have then two partial models of *e* built on *HB*:
>   *j$_{p1}$ ={light(a), white(a)}*
>   *j$_{p2}$ ={light(a), square(a)}*
> The point is that we need information concerning the object *b* in order to reduce the uncertainty about the object *a*: if we only consider information concerning *a*, we have a third partial model of *e* on *HB*, namely *{light(a), white(a), square(a)}* that cannot be the correct description of our example.

Assumptions based learning is a special case of learning from possibilities. More precisely Assumptions based learning reduces to learning from possibilities in such a way that labels of examples are preserved by the data transformation. Let us consider an uncertain extended example *e*, the corresponding uncertain-generalized example $e_p = \{P_1,...,P_u\}$ is such that each $P_k$ is the clausal theory $ct(j_k)$ corresponding to one partial model $j_k$ of *e* built on *HB*. So $e_p$ describe all possible complete examples intended when defining the uncertain extended example *e*.

In Example 4 the uncertain extended example *e* is thus translated as:
$e_p$ = { $P_1$={light(a) ; white(a); ← square(a)}, $P_2$= {light(a) ; ← white(a); square(a)} }

## 5.1 Assumptions based learning of DNF+.

We investigate here the properties of Assumptions based learning when hypotheses are DNF+ formulas as for instance:
$H = \exists (a_1 \wedge ...a_n) \vee \exists (b_1 \wedge ...b_n) \vee ....$ (all literals in the formula are positive literals).

Definition: Order on interpretations
Consider two interpretations $j_1$ and $j_2$. We will say that $j_1$ is smaller (respectively greater) than $j_2$ iff $j_{1p} \subseteq j_{2p}$ (respectively $j_{1p} \supseteq j_{2p}$).

For our learning purpose we are interested in partial models of *e* on *HB*. Such an interpretation is denoted as:
- a <u>maximal</u> partial model of *e* on *HB*, when no greater interpretations on *HB* are partial models of *e*.
- a <u>minimal</u> partial model of *e* on *HB*, when no smaller interpretations on *HB* are partial models of *e*..

Proposition 5  *Let H be a DNF+ hypothesis and e be a clausal theory representing an extended uncertain example then:*
  a) *If there is no <u>maximal</u> partial model of e on HB that is a model of H, then there is no partial model of e on HB that is a model of H.*
  b) *If there is no <u>minimal</u> partial model of e on HB that is <u>not</u> a model of H, then there is no partial model of e on HB that is <u>not</u> a model of H.*
  (proof is given appendix A.4)
Corollary –
  a) *H is compatible$_A^+$ with a positive extended uncertain example e iff there is a maximal partial model of e on HB that is a model of H.*
  b) *H is compatible$_A^-$ with a negative extended uncertain example e if and only if there is a minimal partial model of e on HB that is <u>not</u> a model of H.*
As a consequence we only need to check maximal partial models of *e* built on *HB* when *e* is a positive example. Also we only need to check minimal partial models of *e* built on *HB* when *e* is a negative example.

Furthermore searching partial models of *e* corresponds to a combinatorial search whose tests can be made efficient if we only need positive assumptions. The following proposition shows that this is the case when considering positive examples:

<u>Proposition 6</u> – *H is compatible$_A^+$ with the extended uncertain positive example e iff there exists an interpretation j built on HB such that:*
   $ct(j_p) \land e \mid\neq_{HB)} \square$   *and j is a model of H*
 (proof is given appendix A.5)

Note that negative assumptions are completely useless here, as checking whether an interpretation *j* is a model of a DNF+ hypothesis is also performed only considering the positive part $j_p$ of the interpretation *j*.

Unfortunately when considering negative examples the corresponding property (i.e. replacing "positive„ by "negative„ and "*j* is a model of *H*„ by "*j* is not a model of *H*„ in Proposition 6 ) is not true. Note however that whenever $e^-$ is a Horn clausal theory, then there is only one minimal model, namely the Herbrand minimal model, whose positive ground atoms are the ones that are logical consequences of *e*.
This means that we can just forget assumptions when considering negative examples and in this case the combinatorial search is useless:

<u>Proposition 7</u> – *Let $e^-$ be a negative extended example represented as a Horn clausal theory, let HBe be the associated Herbrand Base, and let HB $\subseteq$ HBe be the Herbrand base used for learning.*
*A DNF$^+$ hypothesis H is compatible$_A^-$ with $e^-$ iff the projection on HB of the unique least Herbrand model of $e^-$ is not a model of H*
(proof is given in A.6)

### 5.2  Constructing Hypotheses in Assumptions based learning of DNF+.

In our setting uncertain examples do not depend on each other, they are independent observations and the color of *b* in an example  may be freely assumed as *green* with no consequences on truth values of *green(b)* in an other example (note that *b* represents two different entities in $e^+$ and $e^-$). This means that a hypothesis *H* may be checked as compatible with each example (positive or negative) independently. However, note that we have to be cautious when the hypotheses are constructed  piece by piece.

**Constructing hypotheses with set covering methods**

Consider for instance a concept learning algorithm that uses the standard greedy set covering algorithm searching for a DNF$^+$ formula: a first conjunction $h_1$ is selected that covers part of the positive examples and no negative examples, then a second one $h_2$ is selected that covers all the uncovered positive examples and no negative instances, then $H = h_1 \lor h_2$ is output as the solution. However, in our uncertainty setting, we should verify that for each negative example $e^-$ there is at least one model of $e^-$ that makes both $h_1$ and $h_2$ compatible with the negative example.  In other words we should first check compatibility of $h_1$ with $e^-$, then compatibility of $h_1 \lor h_2$ with $e^-$ and so on.

> *Example 5*
> Consider the DNF+ formula *H = square $\lor$ light,* and suppose that $h_1$=*square* is compatible$_A^-$ with a given negative example $e^-$ ( there is a model of $e^-$ in which *square* is false*) and *h₂=light is* also compatible$_A^-$ with *e⁻ (*there is a model of *e⁻* in which *light* is false)

Now, if for all models of *e⁻* either *square* is true or *light* is true, then *H* is not compatible with *e⁻*. This may happen for instance with the following negative example: *e⁻ = {red; square ∨ light}*

Fortunately, when examples are represented as *Horn clausal theory*, partial hypotheses can actually be selected independently:

Proposition 8:
*Let H be a DNF+ formula, h be an existential conjunction, and e be a Horn clausal theory representing an extended uncertain example. Then:*
*If  H and  h are both compatible$_A$ with e then the DNF+ H ∨ h is compatible$_A$ with e.*
(proof is given appendix A.7)

### 5.3  The role of local Knowledge and background Knowledge in assumptions based learning.

Here background knowledge as local knowledge is included in the clausal theory *e* representing our state of knowledge about an uncertain extended example. Let us denote an uncertain extended example as *e* = $F_e$ ∧ $K_e$ where $F_e$ represents the facts concerning the example, i.e. positive and negative ground literals of $HB_e$, and where $K_e$ represents clauses representing either rules or constraints. $K_e$ is divided into the Background Knowledge *B* that holds for all examples, and $L_e$ that represents the local knowledge, i.e. clauses and constraints that specifically hold for the example represented as *e*. For instance in example 4, *F={light(a); Near(a,b)}, B={←Near(x,x); ←square(x) ∧ white(x) ∧ Near(x,y)}* and $L_e = \emptyset$.
Now in the definition of *compatible$_A$* we have the following (rewritten) condition :
...   *"iff* there exists an interpretation *j* built on *HB* such that: *ct(j) ∧ $F_e$ ∧ B ∧ $L_e$ |≠$_{HBe}$ □"* So here knowledge helps to establish the set of interpretations *j* built on *HB* that do not contradict the facts $F_e$ and the knowledge $K_e$. Such an interpretation *j* is so made of facts in *F* built on *HB* plus assumptions, consistent with $F_e$ ∧ $K_e$, on truth-values of the other atoms of *HB*.

Furthermore background and local Knowledge can be used in order to make efficient the computation of these interpretations. The idea is that we can make assumptions on a subset *HBa* of *HBe* different from *HB*: we use then the background and local Knowledge in order to deduce the truth value of ground atoms describing the possible complete examples. This results in the following variant of Assumptions Based Learning :

Let $HB_a$ be the part of $HB_e$ on which assumptions are made. Then *H* is compatible with an example *e* whether:
*There exists* an interpretation *a* on $HB_a$ such that:
  *ct(a) ∧ e|≠$_{HBe}$ □*  and
  *ct(a) ∧   e |=$_{HBe}$ ct( j)* where *j* is an interpretation built on *HB,* and
  *j* is a model of *H,* if *e* is a positive example, or
  *j* is not a model of *H,* if *e* is a negative example.

Deduction should be complete i.e. all partial models *j* of *e* built on *HB* should be deducible from assumptions on *HBa*.

In section §5.4.2 we exemplify this variant on our RNA problem.

## 5.4 RNA secondary structure prediction in assumptions based learning framework

### 5.4.1 RNA secondary structures

RNA *secondary structure* prediction is a molecular biology problem that we have represented in the framework of assumptions based learning from Horn clauses.

A RNA molecule corresponds to a sequence of nucleotides and forms a three-dimensional structure. The RNA secondary structure is a simpler representation characterizing the main biological properties of the molecule and consists in a bi-dimensional conformation of *helices* (regions where the stretch of RNA forms spirals). In most cases the secondary structure is unknown and the RNA sequence is used to predict the secondary structure. In this work we use the fact that a RNA sequence contains occurrences of lexical patterns called *palindromes* from which we can predict a set of possible secondary structures: a helix of the secondary structure always corresponds to a palindrome of the sequence but the converse can be false. So the prediction of the secondary structure from an RNA sequence comes to identifying and describing the subset of palindromes that corresponds to the set of helices.

To describe a possible secondary structure we use the three following structural binary relations (as illustrated Figure 5) defined on pair of palindromes: *Precedes* (P), *Overlaps* (O), *Includes* (I).

(1) $P(p_1, p_2)$

(2) $O(p_1, p_2)$

(3) $I(p_1, p_2)$

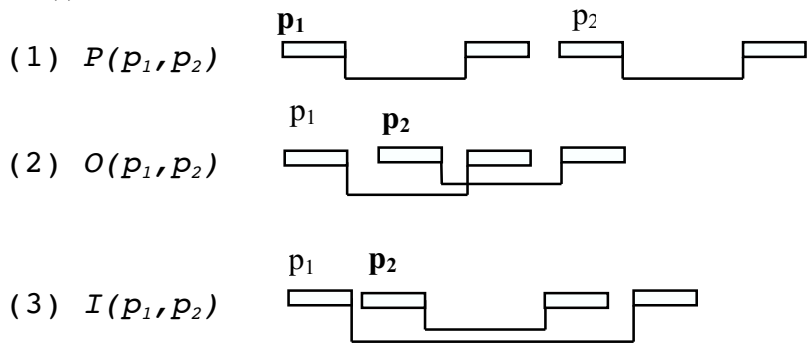

*Figure 5:*(1), (2) and (3) are the three structural relations between two palindromes $p_1$ and $p_2$.

Additional biological information helps to determine the subset of palindromes corresponding to helices and thus representing the structure of a RNA sequence. In particular two palindromes can be structurally *incompatible* and as a consequence cannot both represent helices. Furthermore the set of helices corresponds to one of the maximal subsets of compatible palindromes.

Figure 6 illustrates the set of palindromes of a given sequence, their relations, and two possible secondary structures.

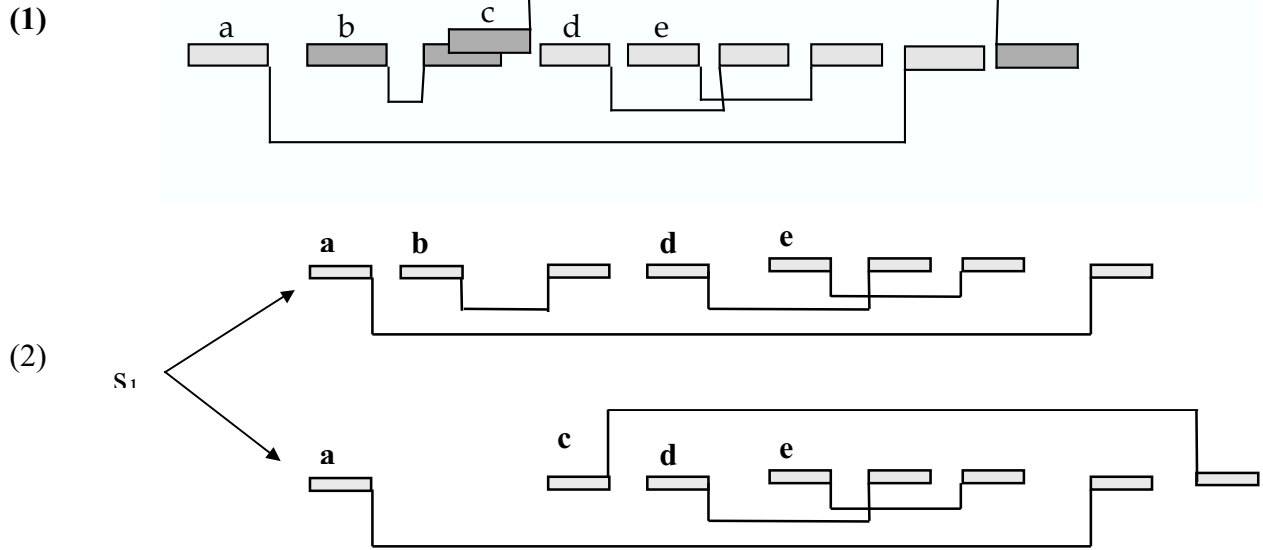

*Figure 6: (1) The palindromes {a, b, c, d}, here b and c are incompatible. (2) The two possible secondary structures*

### 5.4.2 Learning the consensus secondary structure of a set of RNA sequences

The problem we address here consists in learning structural patterns that occur in the structure of the members of a set of RNA molecules. An example is a RNA molecule represented as a sequence of nucleotides. To represent the example we first have to compute the set of palindromes found on the sequence, and the possible structural relations between compatible palindromes together with the pairs of incompatible palindromes.

Here examples are positive uncertain extended examples as found in our assumptions based learning framework: we need incompatibility information to reduce uncertainty, but regarding learning we only need atoms describing structural relations:

- o An example $e$ is then a Horn clausal theory that contains the general constraint of incompatibility of palindromes, structural and incompatibility relations between palindromes. $HB_e$ is the Herbrand Base of this clausal theory.
- o The part $HB$ of $HB_e$ only contains the ground atoms describing the structural relations.

However, as incompatibility concerns palindromes and not structural relations the most natural and efficient way to investigate the various possible structures is first to investigate the maximal subsets of compatible palindromes, and then to deduce the corresponding grounded structural relations. This corresponds to assumptions based learning using background and local knowledge as discussed in §5.3

*Example 7*

> Consider the palindromes given Figure 6.
> The base on which assumptions will be made is:
> $HBa= \{hel(a), hel(b), hel(c), hel(d), hel(e)\}$
> The base $HB$ on which possible complete examples are described is formed from the predicates $O, I, P$ corresponding to the structural relations, and from the constants $\{a,b,c,d,e\}$ representing the palindromes observed in the sequence.

The base *HBe* on which extended examples are represented includes *HB, HBa* and the ground atoms built on the incompatibility relation #.

- #(b,c)← means that the palindromes $p_1$ and $p_2$ are not compatible.
- ← (#(X,Y) ∧ hel(X) ∧ hel(Y)) is the clause defining that two incompatible palindromes cannot be both helices.

When we observe two palindromes $p_1$ and $p2$ on the sequence, we also observe the corresponding structural relation $R(p_1, p_2)$ under the assumption that both $p_1$ and $p_2$ are helices. This corresponds to the clause $R(p_1, p_2)$ ← hel($p_1$) ∧ hel ($p_2$) where R is one of the three structural relations *O, I, P*.

An example *e* = is then represented as the clause that define incompatibility together with the structural information and incompatibilities:
$e = F_e \wedge B \wedge L_e$     where
$F_e$ ={#(b,c)←},
$B$ ={← (#(X,Y ∧ hel (X) ∧ hel(Y))},
$L_e$ =  {I(a,b) ← hel(a) ∧ hel(b); O(a,c) ←hel (a) ∧ hel(c);
      I(a,d) ← hel (a) ∧ hel(d); I(a,e) ← hel (a) ∧ hel(e);
      P(b,d) ← hel(b) ∧ hel(d); P(b,e) ← hel(b) ∧ hel(e) ;
      I(c,d) ← hel(c) ∧ hel(d) ; I(c,e) ←hel (c) ∧ hel(e);
      O(d,e) ←  hel(d) ∧ hel(e)}

We don't represent here negative facts concerning structural relations as ← O(a,b), ←#(b,c), or ←O(a,a), because, given a set of positive assumptions on helices, together with the facts and local and background theory, we deduce the corresponding set of positive assumptions on structural relations that represents a possible structure.
From the two maximal sets of assumptions on *HBa*:
    $a_1$= {hel(a), hel(b), hel(d), hel(e)} and
    $a_2$ ={hel(a), hel(c), hel(d), hel(e)},
we obtain the following possible examples described on *HB*:
    $j_{p1}$ ={I(a,b),  I(a,d), I(a,e), P(b,d), P(b,e), O(d,e)}
    $j_{p2}$ ={O(a,c),  I(a,d), I(a,e), I(c,d), I(c,e), O(d,e)}

A hypothesis corresponds here to a structural pattern. For instance ∃XYZ O(X,Y) ∧I(X, Z)∧I(Y, Z) represents a pattern stating that a helix *X* overlaps an helix *Y* and includes an helix *Z*, and stating that the helix *Y* also includes the helix *Z*.

An issue with Assumptions based learning as described here is that all possibilities have to be considered. Our RNA problem, in consequence, would not be tractable since each RNA sequence we consider actually results in a huge amount of possible examples. Fortunately these possible examples are not equally probable. If we associate to each possible example a probability, then the probability that the example is a model of a given hypothesis *H* is simply the sum of the probabilities of the possible examples that are models of $H$[2]. In our application we use a particular property of our RNA problem: for each sequence we can rank structures and keep only few of the most probable structures with a weak probability to reject the correct

---

[2] Suppose that the possible examples $j_1$ and $j_2$ in Example 7 have probabilities 0.9 and 0.1, then the example is a model of the hypothesis $H_1$ =∃XY O(X,Y) with probability 1, since both $j_1$ and $j_2$ are models of *H*

structure. So, though we had thousands of possible structures for each sequence, only few of them (about 200) were considered. Of course this means that the most probable structures of each sequence are searched only once prior to learning. The overall method and results is extensively described in [Bouthinon and Soldano, 1999b].

# 6 Related work

## 6.1 Multiple instance learning

In this setting originally proposed by Dietterich [Dietterich *et al*, 1997] [3], each example *e* of the target concept is a set $\{inst_1,...inst_n\}$ of descriptions called instances[4]. A positive example $e^+$ is an uncertain example: at least one instance (possibly several ones) satisfies the target concept[5]. A negative example $e^-$ is a generalized example: no instance satisfies the target concept.

Although Dietterich does not actually present this setting in a logical framework we can formalize it in the following way: consider that each instance of an example *e* is an interpretation, then *e* is the formula which models are these interpretations.

The framework is then similar to Learning from satisfiability: positive examples are uncertain, negative ones are generalized. However in multiple-instance problems, the examples (and hypotheses) are expressed in propositional (or attribute-value) representations. Such a learning setting is useful when the learning task concerns multiple realities rather than uncertainty. In the application that motivated multiple instance learning, the concept target was the activity of a drug, each example was a molecule that could take various 3-dimensional conformations, so each instance was one of these conformations. But during the (chemical) experiments most of the conformations of a molecule appear. So in the learning task, an active molecule is such that at least one conformation should be responsible of the activity. However an inactive molecule corresponds to a generalized negative example: none of the conformations should allow the activity. What happens here is that we do not really deal with uncertain or generalized examples. Rather, we have a set of views of one, complete example. The same situation occurs with multiple part problems, as defined by J.D. Zucker [Zucker, 1998], [Zucker and Ganascia 1998], and in various attempts to propositionalize first order learning problems in order to use variants of efficient propositional or attribute-value learners [Alphonse and Rouveirol, 2000], [Sebag *et al*, 1996]. The basic idea is that an object described using a first order representation, i.e. a relational description, can be reduced, starting from some basic pattern, to multiple parts or multiple views corresponding to various matches with a given pattern. As a consequence not only multiple instance or multiple part learning and propositionalization do not deal with uncertainty, but multiple instances as single instances can be uncertain. Note however that in this case, we should take care of dependencies between assumptions.

## 6.2 Hirsh's settings

---

[3] see also [Auer, 1997 ; Maron and Lozano-Perez, 1998]

[5] More precisely a boolean function i is associated with each example *e*: if *e* is positive $\exists inst \in e$ such that *f(inst) = true*, and if *e* is negative $\forall inst \in e, f(inst) = false$.

Extending the work of T. Mitchell [Mitchell, 1982], H. Hirsh introduced Version-Space Merging [Hirsh, 1994] that represents each example *e* (positive or negative) as the set of classifiers (i.e. hypotheses) that are compatible with *e*, (in the same sense used in the present paper), each set being represented as a Version Space. The final Version Space is obtained by intersecting all these Version Spaces. This work uses attribute-value representations of data and hypotheses.

In its paper, H. Hirsh discuss various learning situations that Version Space merging allows to deal with. For instance:
> Ambiguous data: Some attributes of the examples are not perfectly known (they can take one value among a set of possibilities). The approach corresponds to our pure uncertainty setting.
> Explanation-Based Learning: Here the apprentice is supposed to have a background theory that, when applied to complete examples, provides explanations of positive and negative examples, i.e. possibly transforms the original example into one (or more) generalized examples. Whenever considering all these generalized examples as correct, this setting corresponds to our generalized examples setting. However, if the theory is incorrect but making the weaker assumption that for each example at least one of the explanations is correct, then each example results in a set of generalized examples from which at least one is correct (e.g. a bird is light and migratory and we guess that either "*bird* and *light*" or "*bird* and *migratory*" is sufficient to conclude on the target concept). This situation can be addressed in our learning from possibilities setting by considering examples as both uncertain and generalized.

### 6.3   Uncertainty in propositional or attribute-value representations.

Uncertainty in propositional or attribute-value representations is often denoted as the "missing values" problem. There are basically two approaches: either predicting the complete description or taking into account the missing values when scoring the hypotheses. The former approach supposes that one possible identity of the example is much more "probable„ than others and that such accurate prediction is possible. Then simple methods compute a most probable value for each missing value, possibly using some simple Bayesian scheme, where most sophisticated methods attempt to learn from examples to predict the missing values [Liu *et al*, 1997]. In the later approach all possible identities are in some sense considered, as the scoring function to optimize when searching a preferred solution is weighted according to the probability distribution of the possible values for uncertain attributes [Quinlan, 1993].The latter approach could not be used here since in our RNA application all values were missing, so we had to use knowledge about the relations between attribute-values. The former one, selecting only one possible identity was also not adequate (as the most probable RNA structure generally is not the true one).

### 6.4   Abduction and Induction in Inductive Logic Programming.

Regarding first order representations, uncertainty has been addressed in works on abduction and induction (see [Dimopoulos and Kakas, 1996; Kakas and Riguzzi, 1997]). The work of Kakas and collaborators uses the so-called "normal ILP learning settings„ in which there is only one observation, (or database), and examples are grounded predicates that have to be derived from a hypothesis, facts and background knowledge. In such learning settings, knowledge about examples is considered as complete, and following the closed-world assumption, when some grounded predicate is not in the database, it is considered as false. To deal with uncertainty in I_ACL, some predicates are stated as "abducible„, i.e. useful assumptions on which grounded occurrences of these predicates are true can be made in order to derive the examples. Furthermore the scoring function of I-ACL gives a weaker weight to hypotheses that need assumptions to derive examples. Note that, as there is one unique observation and assumptions are made on a set of abducible predicates, the assumptions concerning the examples are then mixed up, resulting in a complex combinatorial problem: If we have $n$ examples each of which requires $k$ (abducible) grounded atoms, then there is $2^{kn}$ sets of assumptions possibly constraining the hypotheses. The same problem in our setting, in which assumptions sets corresponding to various examples are independent, results in $n2^k$ sets of assumptions. This means that choices regarding sets of assumptions have to be made very early during learning in the former setting, and this was not acceptable regarding our biological application. Note, however, that abductive induction settings represent all the examples in one unique world and so allow expressing more dependencies between assumptions.

# 7  Conclusion

We have discussed in this paper logical learning settings dealing with incomplete examples. We have identified two basic kind of incompleteness uncertainty and abstraction and we have investigated a new general setting, namely learning from possibilities, in which each example is associated to a set of possibilities. Each possibility corresponds to an alternative partial description of the example that is stated as sufficient to classify the example with respect to the target concept. This learning setting is shown to extend Learning from satisfiability, which was proposed by L. De Raedt, by allowing more cases of incompleteness. As most learning settings were previously shown to be particular cases of learning from satisfiability, these settings also are particular cases of learning from possibilities. Though learning from satisfiability was defined to learn Clausal theories, we emphasized learning of DNF formulas that are first order extensions of DNF hypotheses as investigated by propositional learners However we have shown that algorithms that learn a DNF hypothesis from possibilities can also learn a Clausal theory by inverting the example labels and negating the resulting DNF hypothesis (and conversely algorithms that learn a Clausal theory can also learn a DNF formula). A particular case of learning from possibilities is assumptions based learning that copes with examples which uncertainty can be reduced when considering contextual information outside of the proper description of the examples, i.e. knowledge about other properties (i.e. predicates) or entities (i.e. constants). In particular we have exhibited various practical properties of assumptions based learning of DNF+ formulas and discussed the use of local and background knowledge both to deduce information about the examples and to help reducing the uncertainty about examples. This was exemplified on our RNA structure prediction problem that led to a first (and ad hoc) implementation of assumptions based learning. A first research direction includes the investigation of using probabilities in assumptions based learning and the design and implementation of efficient general learners for both propositional and first order assumptions based learning. A second research direction concerns assumptions based learning as a model for uncertain relational learning in knowledge discovery and data mining. Further work also addresses investigation of abstraction and uncertainty in the learning from entailment setting, and in learning in hybrid knowledge representations formalisms [Rouveirol and Ventos, 2000]. Finally it would be interesting to investigate the relation between Assumptions Based Learning (ABL) and its deductive counterpart as found in Assumptions Based Truth Maintenance Systems (ATMS) [De Kleer, 1986]. Indeed in ABL a hypothesis is compatible with an uncertain example provided that some assumptions concerning the truth-value of ground atoms are allowed, and, as in ATMS, we have to deal in ABL with the various sets of assumption that support the hypothesis. This suggests that incremental ABL learners could benefit from ATMS techniques.

*Appendix A - Proofs*

In the proofs mentioned hereafter we will consider two Herbrand bases $HB$ and $HBs \subseteq HB$ from which interpretations can be formed and an hypothesis $H = \exists h_1 \vee \exists h_2 ... \vee \exists h_n$ (a $DNF^+$ formula) formed from predicates (and possibly the set of constants) mentioned in $HB$.

A.1: Lemmas

Let us first consider following Lemmas, which will be used in the proofs below.

Lemma 1
Let $j$ be a model of $H$, then an interpretation $i$ greater than $j$ is a model of $H$.
*Proof :* $j$ is a model of $H$ means that there is a substitution $\theta$, grounding at least one $h_k$, such that all grounded atoms of $h_k\theta$ belongs to $j_p$. Then, all grounded atoms of $h_k\theta$ belong to $i_p$ as $i_p \supseteq j_p$.

Lemma 2
Let $j$ be an interpretation that is not a model of $H$, then any interpretation $i$ smaller than $j$ is not a model of $H$.

Lemma 3
Let $j$ be an interpretation built on $HBs$ and $tc(j)$ its associated clausal theory.
An interpretation $j'$ on $HB$ is a model of $tc(j)$ iff $j'$ is an extension of $j$ on $HB$.
*Proof :* Let $j_p = \{t_1, ..,t_m\}$ and $j_n = \{f_1,...,f_m\}$ be the sets of respectively true and false grounded atoms of $HB$ in $j$. . Then $tc(j) = \{t_1, ..,t_m, \leftarrow f_1,..., \leftarrow f_m\}$.
"$\Rightarrow$"Let $j'$ be a model of $tc(j)$.built on $H$. Then $j'$ is a model of each single grounded clause $c$ of $tc(j)$. If $c$ is a positive literal then necessarily $j'_p$ contains $c$. If $c$ is a negative literal $\leftarrow f$, then necessarily $j'_n$ contains $f$. Thus $j'$ assigns to all atoms of $j$ the truth-value they have in $j$, and so $j'$ is an extension of $j$.
"$\Leftarrow$„ As $j'$ is an extension of $j$, then $j'_p \supseteq j_p$ and $j'_n \supseteq j_n$. As a consequence $j'$ is a model of each single grounded clause of $tc(j)$.

A.2: Proof of Proposition 1
Let us consider a learning set under learning from satisfiability $Es = Es^+ \cup E_S^-$. Following De Raedt's notations we will consider that any example of $E_S$ is written as $(c, class)$ where $c$ is the clausal theory that represents this example, and *class* is either *positive* (if $e \in Es^+$) or *negative* (if $e \in Es^-$). The function $\rho$ maps any example *(c, class)* of $Es$ onto an example $\rho((c, class)) = (C, class)$ where $C$ is a set of consistent clausal theories such that:
 – $\rho((c, positive)) = (\{ct(m_1), ...,ct(m_k)\}, positive)$ if we suppose that the clausal theory $c$ has $k$ Herbrand models $\{m_1,...,m_k\}$, $ct(m_i)$ and that $ct(m_i)$ is the clausal theory which unique model is $m_i$.
 – $\rho((c,negative)) = (\{c\}, negative)$

In consequence $\rho$ maps $Es$ onto a set $E_P = E_P^+ \cup E_P^-$. Then we have to prove that:

$Sol(E_S)$ (under *compatibleS*) = $Sol(E_P)$ (under *compatibleP*), where these solution sets are defined as :

$Sol(E_S) = \{H \mid \forall(c, positive) \in E_S^+, H \wedge c \not\models \Box$, and $\forall(c, negative) \in E_S^-\ H \wedge c \models \Box\}$

$Sol(E_P) = \{H \mid \forall(C, positive) \in E_P^+, H$ is $compatible_P^+\ C$, and $\forall(C, negative) \in E_P^-, H$ is $compatible_P^-$ with $C\}$

The proof is as follows:

(1) Let $(c, positive)$ be an example of $E_S^+$ and let $\rho((c, positive)) = (\{ct(m_1), ..., ct(m_k)\}, positive)$ be the corresponding example of $E_P^+$, then :

$H \wedge c \not\models \Box \Leftrightarrow$ at least one Herbrand model of $c$ is a model of $H \Leftrightarrow \exists m_i \in \{m_1,...,m_k\}$ (the Herbrand models of $c$) such that $m_i$ is a model of $H \Leftrightarrow \exists pi = ct(mi) \in \{ct(m_1), ..., ct(m_k)\}$, such that $pi \models H \Leftrightarrow H$ is $compatible_P^+$ with $\{ct(m_1), ...ct(m_k)\}$

(2) Let $(c, negative)$ an example of $E_S^-$ and $\rho((c, negative)) = (\{c\}, negative)$ the corresponding example of $E_P^-$ then

$H \wedge c \models \Box \Leftrightarrow \exists pi = c \in \{c\}$ such that $pi \wedge H \models \Box \Leftrightarrow H$ is $compatible_P^-$ with $\{c\}$

## A.3: Proof of Proposition 2

We provide a counter-example. Let us suppose that $E_P^+$ contains only the following positive example $e_p = \{\{a\},\{b\}\}$ and let $E_P^- = \emptyset$. The concept space contains the clausal theories built from the 0-ary predicates $a$ and $b$. Then $H$ is $compatible_P^+\ e_p$ means that $\exists pi \in \{\{a\},\{b\}\}$ such that $pi \models H$, i.e. *(1)* $a \models H$ or $b \models H$.

Suppose a reduction $\rho$ that maps $E_P = \{(e_p, positive)\}$ onto $E_S$ containing the example $\{(C, class)\}$ where $C$ is a consistent clausal theory and *class* is either *positive* or *negative*. Then a solution $H$ to the corresponding learning from satisfiability problem should either satisfy *(2)* or *(3)* :

*(2)* $C \wedge H \not\models \Box$ if *class = positive*

*(3)* $C \wedge H \models \Box$ if *class = negative*

Let $I = \{\emptyset, \{a\}, \{b\}, \{a, b\}\}$ be the set of all Herbrand interpretations. We denote as *True* the formula whose set of models is $I$.

We first observe that *class* cannot be *negative* as $H = true$ satisfies *(1)* but does not satisfy *(3)* for any consistent clausal theory $C$.

Suppose now that *class = positive*. As neither $H = \neg a$ nor $H = \neg b$ satisfy *(1)*, they must not satisfy *(2)* and in consequence $C \models a$ and $C \models b$. We conclude that $C$ has one single model that is $\{a,b\}$. We observe that for such a formula $C$, $H = a \wedge b$ satisfies (2) but does not satisfy (1) and this contradicts our initial assumption.

## A.4: Proof of Proposition 5

- Proposition 5-a
  Consider a partial model of $e$ which is model of $H$, then, following Lemma 1, there exists a maximal partial model of $e$ which is a model of $H$. Conversely if each maximal partial model of $e$ is not a model of $H$ then, following Lemma 2, no partial model of $e$ is a model of $H$.

- Proposition 5-b
  Consider a model of $e$ that is not a model of $H$, then following Lemma 2, there exists a minimal partial model of $e$ that is not a model of $H$. Conversely if each minimal partial model of $e$ is a model of $H$ then, according to Lemma 1, each partial model of $e$ is a model of $H$.

The corollaries *a)* and *b)* are obvious.

## A.5: Proof of Proposition 6

$\Rightarrow$ : $H$ is compatible$_A^+$ with $e$ means that there is a partial model $k$ of $e$ built on $HB$ and such that $ct(k_p) \wedge ct(k_n) \wedge e| \neq \Box$. Clearly $k$ is also such that $ct(k_p) \wedge e| \neq \Box$.

$\Leftarrow$ : Let us consider a model $k$ of $ct(j_p) \wedge e$ built on $HB$. Then $k$ can be written as $m + n$, where $m$ contains the atoms belonging to $HB$ and $n$ contains the atoms belonging to $HBe-HB$. Then $m = m_p + m_n$, and as $k$ is a model of $ct(j_p)$, then $m_p \supseteq j_p$.

The interpretation $m$ built on $HB$ is so such that :
- $m$ is a model of $H$ (because $m$ is greater than $j$ and $j$ is a model of $H$ (see Lemma 1))
- $ct(m_p) \wedge ct(m_n) \wedge e| \neq \Box$ (because $m$ is included in $k$ which is a model of $e$).

## A.6: Proof of Proposition 7

Let $m$ be the least Herbrand model of $e^-$, and $pm$ be its projection on $HB$.

"$\Leftarrow$" Suppose $pm$ is not a model of $H$, then $H$ is compatible with $e^-$ if $ct(pm) \wedge e^-|\neq_{HBe} \Box$. According to Lemma 3 ("$\Leftarrow$„) $m$ is a model of $ct(pm)$. As $m$ is also a model of $e^-$ then $H$ is compatible with $e^-$.

"$\Rightarrow$" Suppose that $H$ is compatible with $e^-$ thanks to an interpretation $j$ of $HB$. Then $j$ is not a model of $H$ and $ct(j)$ has a model $j'$ on $HBe$ which is a model of $e^-$. According to Lemma 3 ("$\Rightarrow$„), $j'$ is an extension of $j$.

Consider now the following inclusions :

- $pm_p \subseteq j'_p$, because $pm_p \subseteq m_p$ (as $m$ is an extension of $pm$) and $m_p \subseteq j'_p$ (as $j'$ is greater than $m$, like all models of $e^-$)
- $j_p \subseteq j'_p$ (as $j'$ is an extension of $j$)

As $j$ determines the truth-value of all atoms of $HB$, then $j'_p - j_p$ is only made of true atoms of $HBe-HB$. Suppose now that $pm$ is a model of $H$ on $HB$. In such a case as $pm_p \subseteq j'_p$, then $pm_p \subseteq j_p$. Actually each atom of $pm_p$ is in $HB$ so it cannot be in $j'_p - j'$. Then, according to Lemma 1 $j$ is a model of $H$, which contradicts the initial assumption that $j$ is not a model of $H$. In consequence $pm_p$ (actually $pm$) is not a model of $H$.

## A.7 : Proof of proposition 8

Let $e^+$ be a Horn clausal theory representing a positive extended uncertain example. $H$ is compatible with $e^+$ means that there is a model $k$ of $H$ built on $HB$ and such that $tc(k)$ has a model built on $HBe$ that is a model of $e^+$. According the laws of logics $k$ is also a model of $H \vee h$, so $H \vee h$ is compatible with $e^+$.

Let now $e^-$ be a Horn clausal theory representing a negative extended uncertain example. $e^-$ has a unique least Herbrand model $m$ on $HBe$, and we will call $pm$ its projection on $HB$.

As demonstrated in proposition 7, $H$ is compatible with $e^-$ entails that $pm$ is not a model of $H$. As $h$ is compatible with $e^-$, for the same reason $pm$ is not a model of $h$.

So we observe that $H \vee h$ is compatible with $e^-$ :
a) $pm$ is not a model of $H \vee h$ ($pm$ would then be a model of $H \vee h$ such that $pm$ was a model of $H$ or $pm$ was a model of $h$, which as previously been proved as false).
b) $ct(pm)$ is a model of $e^-$ : $m$ is the extension of $pm$ in $HBe$ and therefore is a model of $ct(pm)$ built on $HBe$ (see Lemma 3 ("$\Leftarrow$")).